\documentclass[10pt,twocolumn,letterpaper]{article}

\usepackage[pagenumbers]{iccv}      

\definecolor{iccvblue}{rgb}{0.21,0.49,0.74}
\usepackage[pagebackref,breaklinks,colorlinks,allcolors=iccvblue]{hyperref}

\usepackage{cuted}
\usepackage{multirow} 
\renewcommand{\thefootnote}{}

\def\paperID{13106} 
\def\confName{ICCV}
\def\confYear{2025}

\title{FlexIP: Dynamic Control of Preservation and Personality for Customized Image Generation}

\author{
        Linyan Huang$^{\dagger}$  
\quad   Haonan Lin$^{\dagger}$
\quad   Yanning Zhou
\quad   Kaiwen Xiao \vspace{10px}
\\ 
Tencent AIPD
\vspace{-30px}
}

\begin{document}
\maketitle

\def\thefootnote{$\dagger$}\footnotetext{Equal contribution.}\def\thefootnote{\arabic{footnote}}

\begin{strip}
    \centering
    \includegraphics[width=\textwidth]{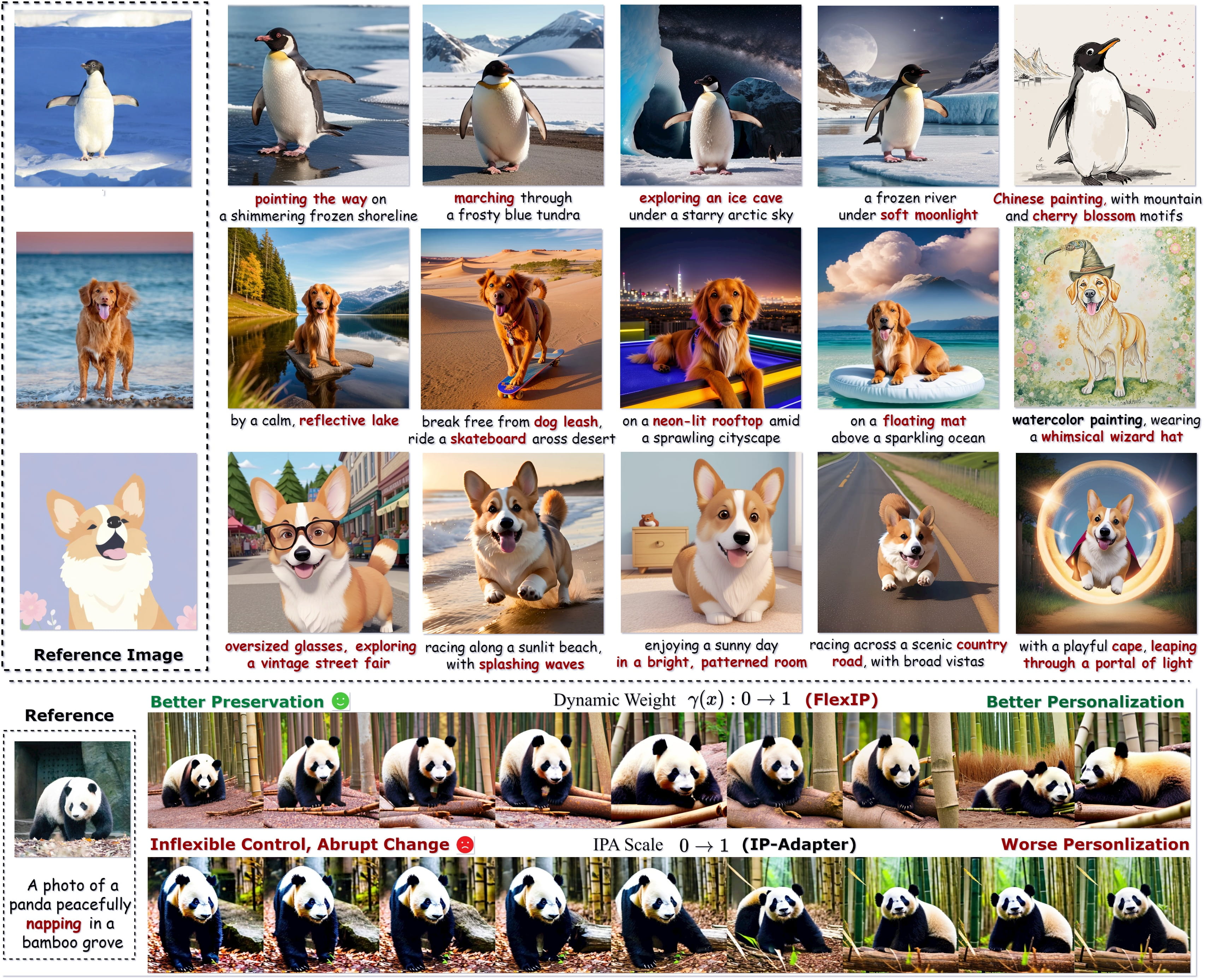} 
    \captionof{figure}{
    \textbf{Top}: \textbf{FlexIP} showcases versatility and precision in personalized image generation. Given a single reference image (left column), it vividly captures identity details while creatively following diverse text prompts, resulting in coherent yet highly varied edits.
    \textbf{Bottom}: \textbf{FlexIP}'s dynamic weight gating mechanism smoothly transitions between strong identity preservation and diverse personalization, significantly outperforming IP-Adapter, which suffers from abrupt identity shifts and rigid control. This reflects superior flexibility and user-friendly controllability.}
    \label{fig:teaser}
\end{strip}

\begin{abstract}


With the rapid advancement of 2D generative models, preserving subject identity while enabling diverse editing has emerged as a critical research focus. Existing methods typically face inherent trade-offs between identity preservation and personalized manipulation. We introduce \textbf{\textit{FlexIP}}, a novel framework that decouples these objectives through two dedicated components: a Personalization Adapter for stylistic manipulation and a Preservation Adapter for identity maintenance. By explicitly injecting both control mechanisms into the generative model, our framework enables flexible parameterized control during inference through dynamic tuning of the weight adapter. Experimental results demonstrate that our approach breaks through the performance limitations of conventional methods, achieving superior identity preservation while supporting more diverse personalized generation capabilities (\href{https://flexip-tech.github.io/flexip/}{Project Page}).

\end{abstract}
\section{Introduction}

The swift progress of 2D diffusion models~\cite{rombach2022high, zhang2023controlnet} has propelled ongoing advancements in image synthesis~\cite{podell2024sdxl} and editing technologies~\cite{brooks2023instructpix2pix}. These models demonstrate remarkable abilities to generate high-quality and diverse visual content from textual or visual input, showing immense potential in artistic creation and advertising design.


Current research in subject-driven image generation primarily follows two paradigms: 
inference-time fine-tuning and zero-shot image-based customization.
The fine-tuning approach~\cite{ruiz2023dreambooth, gal2022image, ruiz2024hyperdreambooth} learns pseudo-words as compact subject representations, requiring per-subject optimization. While this achieves high-fidelity reconstruction, it inherently sacrifices editing flexibility due to overfitting on narrow feature manifolds. 
In contrast, zero-shot methods~\cite{chen2024anydoor, zhang2024attention, nam2024dreammatcher} leverage cross-modal alignment modules trained without subject-specific fine-tuning, achieving greater editing flexibility but often compromising identity integrity. 
Fundamentally, existing methods treat identity preservation and editing personalization as inherently conflicting objectives, forcing models to make implicit trade-offs. 

We identify a critical limitation in existing zero-shot methods: they often adopt alignment modules similar to the Q-former~\cite{alayrac2022flamingo, li2023blip} from vision-language models (VLMs) to align image-text modalities. While effective in visual understanding for text generation, such modules become insufficient for image generation tasks, as they require capturing broader and more complex visual knowledge. This image-text misalignment results in identity preservation and editorial fidelity not working harmoniously together. Therefore, a more explicit decomposition of visual and textual information is necessary—assigning images to handle identity preservation and texts to guide personalization instructions. Separating these information flows enables each modality to specialize, fostering stronger complementarity and achieving superior cross-modal alignment. 

%
To address these issues, we propose \textbf{\textit{FlexIP}}, the first framework to explicitly decouple identity preservation and personalized editing into independently controllable dimensions. Inspired by the principle of \textit{"low coupling, high cohesion,"} we introduce a \textbf{dual-adapter architecture}, enabling each adapter to focus clearly and independently on its specific task—identity preservation or personalized editing—thus maximizing their complementary strengths.
Specifically, the \textbf{preservation adapter} captures essential identity details by retrieving both high-level semantic concepts and low-level spatial details through cross-attention layers. Intuitively, this approach resembles recognizing a person not just by general descriptors (\textit{e.g.}, age or contour) but also by detailed visual cues (\textit{e.g.}, facial features or hairstyle), thereby robustly preserving their identity even under diverse edits. 
On the other hand, the \textbf{personalization adapter} interacts with the text instructions and high-level semantic concepts. The text instructions provide editing flexibility, while the high-level semantic concepts ensure identity preservation.
By separating identity and personalization feature flows, our design eliminates feature competition found in traditional single-path approaches, enabling explicit decoupling of ``\textit{what to preserve}" and ``\textit{how to edit}." 

As illustrated in Fig.~\ref{fig:teaser} bottom, by changing preservation scale, existing methods produce abrupt transitions between identity preservation and personalization, making precise control challenging. 
Motivated by this, we aim to achieve an explicit control between identity preservation and personalization, and thus introduce a \textbf{dynamic weight gating mechanism} that interpolates between two complementary adapters during inference. Users can continuously adjust adapter contributions, flexibly balancing preservation and personalization (Fig.~\ref{fig:teaser} bottom).
Our empirical analysis reveals a critical dependency between training data modality and adapter efficacy: video-frame training pairs inherently capture temporal deformations (\textit{e.g.}, pose variations, lighting changes), enabling flexible feature disentanglement, whereas static image pairs tend to induce copy-paste artifacts due to overfitting on rigid spatial correlations. To mitigate this, we implement a modality-aware weighting strategy:
preservation adapter dominance (higher 
preservation weight) for image-trained scenarios, enforcing strict identity consistency through feature locking in cross-attention maps. Personalization adapter dominance (higher personalization style) for video-trained scenarios, leveraging temporal coherence to guide structurally coherent deformations. The adapters govern distinct aspects of the generation process:
This dynamic weight gating mechanism transforms the traditionally binary preservation-edit trade-off into a continuous parametric control surface. This enables applications ranging from nuanced, identity-consistent retouching to radical yet coherent subject transmutation.

Our contributions are threefold: First, we introduce FlexIP, a novel plug-and-play framework that decouples identity preservation and personalized editing into independently controllable dimensions, addressing the inherent trade-offs in existing methods. Second, we propose a dual-adapter architecture comprising a preservation adapter and a personalization adapter, which respectively handle identity-critical features and editing flexibility, thereby eliminating feature competition and enhancing edit fidelity.  Third, we develop a dynamic weight gating mechanism that allows for continuous modulation between identity preservation and personalization. Our extensive experiments demonstrate that FlexIP significantly improves identity preservation accuracy while maintaining high levels of editing flexibility, outperforming state-of-the-art methods.


\section{Related Work}


\subsection{Subject-driven Image Generation}
Recent advances in customized image generation primarily follow two paradigms: tuning-based and tuning-free methods. Methods like Textual Inversion~\cite{gal2022image}, DreamBooth~\cite{ruiz2023dreambooth}, and DreamTuner~\cite{hua2023dreamtuner} learn target concepts by fine-tuning a pretrained text-to-image model with a specialized token or prompt. While these approaches~\cite{hua2023dreamtuner, kumari2023multi} achieve strong identity preservation through direct parameter optimization, they suffer from prohibitive computational overhead from per-subject optimization, reduced editability due to overfitting on narrow concept distributions and inherent latency in serving novel concepts. 
To address these limitations, recent works~\cite{li2023blipdiffusion, chen2024anydoor, chen2023photoverse, shi2024instantbooth, ssrencoder} employ pretrained visual encoders to bypass test-time fine-tuning. BLIPDiffusion~\cite{li2023blipdiffusion} aligns image-text pairs via BLIP-2’s~\cite{li2023blip2} cross-modal attention for zero-shot adaptation but struggles with disentangling subject identity from contextual attributes.
IP-Adapter~\cite{ye2023ipadapter} and InstantID~\cite{wang2024instantid} inject identity features via cross-attention modulation, though their static fusion mechanisms lead to implicit entanglement of preservation and stylization objectives.
MSDiffusion~\cite{msdiffusion} constrains edits via spatial attention maps, sacrificing free-form stylization for geometric consistency.
CustomContrast~\cite{chen2024customcontrast} use contrastive learning to decouple subject intrinsic attributes from irrelevant attributes. But it still conduct the implict trade-off between preservation and personalization, which hinders the further improvement of the model. DisEnvisioner~\cite{he2024disenvisioner} extract the subject-essential features while filtering out irrelevant information but inherits the copy-paste artifact from static image training pairs. In this paper, we enable explicit control over the trade-off between
identity preservation and stylistic personalization, allows users to continuously balancing feature rigidity and editability.

\begin{figure}[ht]
    \centering
    \includegraphics[width=0.95\linewidth]{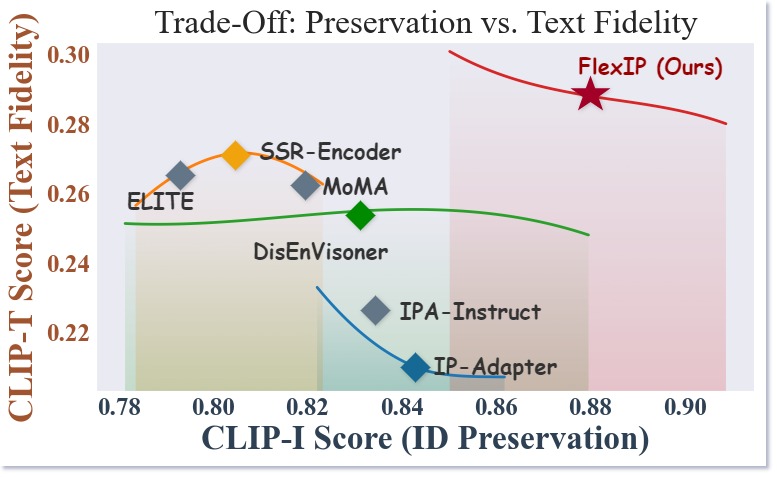}
    \caption{Comparison with other methods on two indicators, image preservation and text fidelity, demonstrates that our approach surpasses previous methods in both aspects.}
    \label{fig:preserve-analysis}
\end{figure}

\subsection{Preservation-Personalization Trade-off}

A core challenge in personalized image generation is balancing identity preservation against editing flexibility, typically measured through text fidelity (alignment with textual instructions). As illustrated in Fig.~\ref{fig:preserve-analysis}, 
existing methods~\cite{wei2023elite, ye2023ipadapter, rowles2024ipadapter, he2024disenvisioner, ssrencoder, song2024moma} show an inherent compromise: methods optimized for high identity preservation (high CLIP-I scores) generally exhibit reduced text fidelity, while those achieving greater editing freedom frequently sacrifice identity consistency.
This trade-off arises due to conflicting optimization goals: strong identity preservation demands strict adherence to reference features, constraining editability, whereas flexible edits encourage semantic diversity at the risk of drifting from the original identity. Thus, we ask: \textbf{Can a method simultaneously achieve robust identity preservation and faithful textual controllability for personalization}?

To address this critical question, our proposed framework, FlexIP, explicitly decouples identity preservation from personalization. By introducing independent adapters controlled through a dynamic weight gating mechanism, FlexIP navigates this trade-off more effectively. This design allows continuous, precise balancing of feature rigidity (preservation) and editability (personalization), which is detailed in the following.



\section{Method}
In this section, we begin by providing a foundational overview of text-to-image diffusion models, including their core mechanisms and relevance to our work. Building on this basis, we present a comprehensive exposition of the proposed FlexIP framework. Specifically, we first elucidate the key observations and challenges that motivated its development, followed by a systematic breakdown of its architecture and operational workflow, detailing its innovative methodology for enabling subject preservation and personalization using a pre-trained text-to-image diffusion model.

\begin{figure*}
    \centering
    \includegraphics[width=0.95\linewidth]{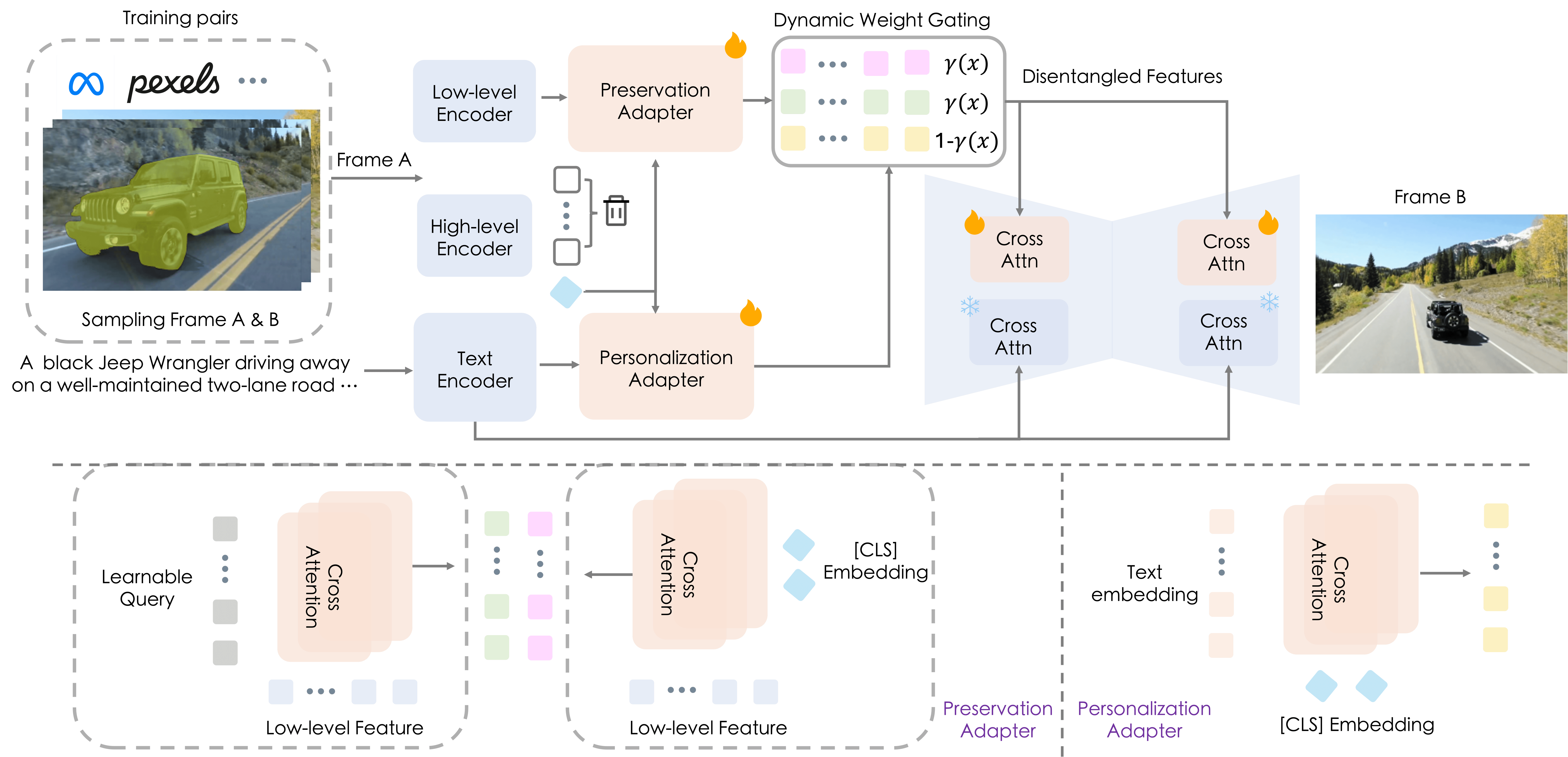}
    \caption{\textbf{The overall pipeline of FlexIP.} It introduces three key improvements to the model: the preservation adapter, the personalization adapter, and dynamic weight gating. First, the preservation adapter combines high-level and low-level features to ensure preservation. The personalization adapter interacts with text and visual CLS tokens to absorb meaningful visual cues, grounding textual modifications within a coherent visual context. Finally, dynamic weight gating navigates the trade-off between personalization and preservation more effectively through independent adapters controlled by a dynamic weight gating mechanism.}
    \label{fig:pipeline}
\end{figure*}

\subsection{Preliminaries}
\textbf{Diffusion Models.}
Diffusion models~\cite{ho2020denoising, song2021denoising} are generative models consisting of two core processes: (i) a forward (diffusion) process that gradually adds Gaussian noise to data, and (ii) a denoising process, guided by conditions such as text prompts, to reconstruct images from noise. The training objective of the noise prediction network $\epsilon_\theta$ optimizes:
\[
\mathcal{L}_{\text{simple}} = \mathbb{E}_{x_0, \epsilon \sim \mathcal{N}(0, \mathbf{I}), c, t} \left| \epsilon - \epsilon_\theta(x_t, c, t) \right|^2,
\]
where \( x_0 \) is clean data, \( c \) denotes conditioning signals, and $x_t$ represents the noisy state at timestep $t$.

Classifier-free guidance~\cite{ho2021classifierfree} enhances conditional generation by training the model with randomly dropped conditioning signals. During inference, predictions blend conditional and unconditional outputs:
\[
\hat{\epsilon}_\theta(x_t, c, t) = w \cdot \epsilon_\theta(x_t, c, t) + (1 - w) \cdot \epsilon_\theta(x_t, t),
\]
where \( w > 1 \) controls the strength of conditioning.
Our work is built upon Latent Diffusion Model~\cite{rombach2022high}, which is conditioned on text embeddings from a CLIP text encoder. 

\textbf{Resampler with Unified Input Representations.}
The Resampler serves as a bridge, connecting \textit{input queries}—designed to capture refined identity information—with retrieval embeddings that store rich, albeit sparse, visual details:
\begin{equation}
    \label{eq:resampler}
    \mathbf{Z}^{(R, X)} = \text{Resampler} \Big( \mathbf{Z}^{(X)}, \mathbf{Z}^{(D)} \Big),
\end{equation} 
where \(\mathbf{Z}^{(X)}\) denotes the input query, and \(\mathbf{Z}^{(D)}\) denotes retrieval embedding.  

The input queries originate from three types of embeddings, all mapped into $\mathbb{R}^d$:
\begin{itemize}[leftmargin=.3in]
    \item Learnable Query Embeddings: $\mathbf{Z}^{(L)} \in \mathbb{R}^{N_L \times d}$,
    \item CLIP [CLS] Embeddings: $\mathbf{Z}^{(C)} \in \mathbb{R}^{N_C \times d}$,
    \item CLIP Text Embeddings: $\mathbf{Z}^{(T)} \in \mathbb{R}^{N_T \times d}$.
\end{itemize}

The retrieval embeddings are derived from DINO Patch Embeddings, which are also mapped into the shared latent space \(\mathbf{Z}^{(D)} \in \mathbb{R}^{N_D \times d}\), capturing detailed visual information from reference images.
By leveraging several perceiver cross-attention (PSA) layers~\cite{jaegle2021perceiver}, Resampler ensures that input queries effectively extract identity-relevant features from these retrieval embeddings:
\begin{align*}
    & \mathbf{A} =  \text{softmax}\left(\frac{\mathbf{Q}(\mathbf{Z}^{(X)}) \mathbf{K}(\mathbf{Z}^{(X)} \oplus \mathbf{Z}^{(D)})^\top}{\sqrt{d}}\right), \\
    & \text{PSA}^{\text{out}} = \mathbf{A} \mathbf{V}(\mathbf{Z}^{(X)} \oplus \mathbf{Z}^{(D)})
\end{align*}
where \(\oplus\) denotes concatenation. In this way, the output \(\mathbf{Z}^{(R, X)}\) integrates rich, low-level visual details from DINO embeddings with the query's high-level semantic context. This refined embedding serves as effective identity conditioning for the subsequent diffusion generation steps.

\subsection{Preservation Adapter}
The first step in ensuring identity preservation is determining which queries and which features should be used to retrieve subject-specific attributes. That is, what kind of queries can effectively extract identity-rich information?

\textbf{Learnable queries for adapatability.}  
To generalize across different subjects, an intuitive approach is to learn representations directly from the data distribution. Unlike static embeddings, learnable queries \(\mathbf{Z}^{(L)}\) provide a trainable subject representation that dynamically adapts to diverse subjects. These queries form a flexible latent space, capable of encoding subject-specific details while remaining generalizable across different styles and conditions.

\textbf{[CLS] embedding for global identity representation.} 
Moreover, CLIP [CLS] embeddings \(\mathbf{Z}^{(C)}\) serve as a pre-trained holistic identity descriptor, which encapsulate high-level semantics such as structure, style in a compact form, offering stability and robustness in identity preservation.

\textbf{Why do these two complete each other?} 
Preserving both fine-grained and global identity attributes is often treated as a trivial challenge.
However, as shown in \textit{Appendix 1.1}, we found that learnable queries specialize in capturing fine-grained variations but lack strong global coherence, while CLIP [CLS] embeddings provide global identity consistency but may miss subtle subject details.  
Therefore, instead of relying on a single embedding to learn both, we adopt a ``divide and conquer" strategy that integrating both for retrieving fine-grained adaptability and global robustness simultaneously from DINO patch embeddings (as shown in Fig.~\ref{fig:pipeline} left bottom), ensuring that identity preservation remains stable even during edits.

Formally, we independently resample the learnable queries \(\mathbf{Z}^{(L)}\) and CLIP [CLS] embedding \(\mathbf{Z}^{(C)}\) through cross-attention with DINO patch embeddings \(\mathbf{Z}^{(D)}\):
\begin{gather}
    \label{eq:preserve_embeds}
    \vspace{-10pt}
    \begin{split}
        \mathbf{Z}^{(R, L)} &= \text{Resampler}_{L}\left(\mathbf{Z}^{(L)}, \mathbf{Z}^{(D)}\right),\\
        \mathbf{Z}^{(R, C)} &= \text{Resampler}_{C}\left(\mathbf{Z}^{(C)}, \mathbf{Z}^{(D)}\right),\\
        \mathbf{P} &= \mathbf{Z}^{(R, L)} \oplus \mathbf{Z}^{(R, C)},
    \end{split}   
\end{gather}
where \(\oplus\) denotes concatenation.  
And \(\mathbf{P}\) serves as identity preservation, which integrates fine-grained local details (via learnable queries) and global semantics (via CLIP [CLS]).

\subsection{Personalization Adapter}
Considering personalization, Stable Diffusion already condition UNet latents on textual embeddings through cross-attention. However, this conditioning provides only general semantic guidance and lacks explicit grounding in the subject's specific visual identity. Consequently, relying solely on the original textual embeddings can cause misalignment between the intended edits and the subject’s appearance.

We address this limitation by introducing an additional personalization adapter, where textual embeddings explicitly attend to the CLIP [CLS] embedding. This additional resampling step enables text embeddings to absorb meaningful visual cues, grounding textual modifications within a coherent visual context. As a result, the textual instructions become more identity-aware, guiding edits that are both accurate and consistent with the subject’s appearance.

Formally, the personalization adapter functions as:
\begin{align}
\mathbf{S} = \mathbf{Z}^{(\text{R}, T)} &= \text{Resampler}_{T}\left(\mathbf{Z}^{(T)}, \mathbf{Z}^{(C)}\right),
\end{align}
where \( \mathbf{Z}^{(T)} \in \mathbb{R}^{N_T \times d} \) are text embeddings (queries), and \(\mathbf{Z}^{(C)} \in \mathbb{R}^{1\times d}\) are CLIP [CLS] embeddings (key-value pairs). Through this, textual guidance is no longer isolated; instead, it becomes visually contextualized, resulting in more precise, flexible and identity-consistent edits.

\subsection{Dynamic Weight Gating}
To address the inherent trade-off between preservation capability and stylized freedom in existing methods, we propose a novel dynamic weight gating (DWG) mechanism for joint training on image and video datasets. Empirical analysis reveals that image data enhances preservation quality but induces copy-paste artifacts~\cite{chen2024anydoor} and weakens instruction adherence, while video data promotes temporal diversity but compromises preservation strength. Our framework leverages the complementary strengths of both modalities by dynamically adjusting the contributions of two specialized adapters. Preservation adapter $\mathbf{P}$ optimized to maintain high-fidelity details and instruction consistency from image data. Personalization adapter $\mathbf{S}$ designed to inject temporal diversity and stylized freedom from video data. 

\begin{table*}[ht]
\centering
\begin{tabular}{l | c c cc cc | cc}
\toprule
\multirow{2}{*}{\textbf{Method}} 
& \multirow{2}{*}{\textbf{mRank}} 
& \multicolumn{1}{c}{\textbf{Personalization}} 
& \multicolumn{2}{c}{\textbf{Preservation}} 
& \multicolumn{2}{c}{\textbf{Image Quality}}
& \multicolumn{2}{|c}{\textbf{User Study (\%)}} \\
\cmidrule(lr){3-3}\cmidrule(lr){4-5}\cmidrule(lr){6-7}\cmidrule(lr){8-9}
 &  & \textbf{CLIP-T} & \textbf{CLIP-I} & \textbf{DINO-I} & \textbf{CLIP-IQA} & \textbf{Aesthetic} & \textbf{Flex} & \textbf{ID-Pres} \\
\midrule
BLIP-Diffusion~\cite{li2023blipdiffusion}                       
            &  8.8  &   0.166 &    0.681 &    0.374 &    0.486 &    5.234   & --- & ---\\
ELITE~\cite{wei2023elite} 
            &  6.2  &    0.269 &    0.793 &    0.657 &    0.522 &    5.437  & --- & ---\\
MoMA~\cite{song2024moma}                                 
            &  5.8  &   0.265 &    0.830 &    0.656 &    0.546 &    5.437   &  9.43 & 7.26 \\
SSR-Encoder~\cite{ssrencoder}                 
            &  5.2  &   0.277 &    0.802 &    0.581 &    0.568 &    5.578   &  6.67 & 3.28 \\
IP-Adapter~\cite{ye2023ipadapter}                           
            &  4.2  &   0.209 &    \underline{0.855} &    0.728 &    0.581 &    5.594 & 4.33 & 2.23 \\
IP-Adapter-Instruct~\cite{rowles2024ipadapter}        
            &  4.8  &   0.234 &    0.833 &    0.701 &    0.584 &    5.459   & --- & --- \\
\(\lambda\)-Eclipse~\cite{patel2024lambda}        
            &  4.4  &   \textbf{0.296} &    0.747 &    0.467 &    \underline{0.589} &    5.597  & 12.5 & 6.97 \\
DisEnVisoner~\cite{he2024disenvisioner}
            &  4.4  &   0.255 &    0.851 &    \underline{0.732} &    0.470 &    \underline{5.658} & 5.67 & 3.52  \\

\midrule
\textbf{\textit{FlexIP (Ours)}}                            
            &  \textbf{1.2}  &   \underline{0.284} &    \textbf{0.873} &    \textbf{0.739} &    \textbf{0.598} &    \textbf{6.039} & \textbf{61.4} & \textbf{76.8} \\
\bottomrule
\end{tabular}
\caption{Comparison of different methods, reorganized by controllability (CLIP-T, Image Reward), similarity (CLIP-I, DINO-I), and image quality (CLIP-IQA, Aesthetic). "Flex" denotes the model's controllability, allowing for adjustable and dynamic modifications. "ID-Pres" represents the model's ability to preserve the identity of the reference image. Bold text indicates the best result, while underlined text denotes the second-best result.}
\label{tab:comparison}
\end{table*}

The DWG mechanism adaptively reweights $\mathbf{P}$ and $\mathbf{S}$ based on the input data type. Let $x \in \mathcal{D}_{\text{img}} \cup \mathcal{D}_{\text{vid}}$ denote a training sample from either the image $\mathcal{D}_{\text{img}}$ or video $\mathcal{D}_{\text{vid}}$ dataset. The feature representation $\mathbf{h}(x)$ is computed as a gated fusion:

\begin{equation}
    \mathbf{h}(x) = \gamma(x) \cdot \mathbf{P} + (1 - \gamma(x)) \cdot \mathbf{S},
\end{equation}
where \( \gamma(x) \) is a data-dependent gating weight given by:

\begin{equation}
    \gamma(x) = 
    \begin{cases} 
        \alpha, & \text{if } x \in \mathcal{D}_{\text{img}}, \\
        1 - \beta, & \text{if } x \in \mathcal{D}_{\text{vid}},
    \end{cases}
\end{equation}
here, $\alpha \in [0, 1]$  and $\beta \in [0, 1]$  are  parameters initialized to prioritize $\mathbf{P}$ for images ($\alpha \rightarrow 1$) and $\mathbf{S}$ for videos ($\beta \rightarrow 1$). This formulation ensures:
image-centric training amplifies P to maximize preservation, ensuring that the essential features of the image are retained. In contrast, video-centric training suppresses
\(\mathbf{P}\) to enhance the stylization capabilities of
\(\mathbf{S}\), allowing for more dynamic and expressive transformations that are suited to video data.
This adaptive mechanism enables the model to dynamically balance preservation and stylization without relying on manual heuristics, effectively leveraging the strengths of both data modalities. By transforming the traditionally binary preservation-edit trade-off into a continuous parametric control surface, this approach could facilitate a wide range of applications.



\section{Experiments}

\subsection{Training Dataset}

Training ideally requires image pairs showing the same subject in varied scenes or viewpoints, but such data are typically unavailable. Previous methods~\cite{ssrencoder, ye2023ipadapter} rely on simple augmentations that fail to represent realistic pose and viewpoint variations.
We follow previous works~\cite{chen2024anydoor,mimicbrush} by utilizing multi-view and video datasets, which naturally provide multiple frames of the same subject. 


Our dataset includes 1.23M varied samples and 11M 
invariant images, covering facial images, natural scenes, virtual try-on, human actions, saliency, and multi-view objects. To balance diversity and generalization, we resample video data to maintain a 1:1 ratio between invariant and varied data, avoiding redundancy. For more details on the dataset construction, please refer to the supplementary materials.

Moreover, previous works often use simplistic and uniform textual prompts across video frames, limiting the model's ability to follow nuanced instructions. To improve textual conditioning and editing flexibility, we use Qwen2-VL~\cite{yang2024qwen2} to generate high-quality, distinct captions for each frame. This approach enhances the diversity and semantic relevance of textual guidance, improving the model's ability to follow detailed editing instructions.

\subsection{Evaluation Dataset and Metrics}
We collect evaluation data from DreamBench+~\cite{peng2024dreambench++} and MSBench~\cite{msdiffusion}, comprising 187 unique subjects. Each image is tested using its set of 9 prompts, with 10 generation runs per prompt. This procedure results in 16,830 customized images used for comprehensive evaluation.

We assess our model using several metrics. For identity preservation, we calculate similarity scores using DINO-I~\cite{zhang2022dino}, CLIP-I~\cite{radford2021learning}, after applying segmentation~\cite{liu2023groundingdino, ren2024groundedsam} to remove background interference. For personalization, CLIP-T measures the semantic alignment between generated images and prompts in the CLIP text-image embedding space.
Moreover, image quality is assessed using CLIP-IQA~\cite{clip-iqa} and CLIP-Aesthetic scores~\cite{aesthetic-score}. Additionally, we compute the mean ranking (mRank) of all metrics for each method to provide an overview of its overall performance.

\begin{figure*}[ht]
    \centering
    \includegraphics[width=1.0\linewidth]{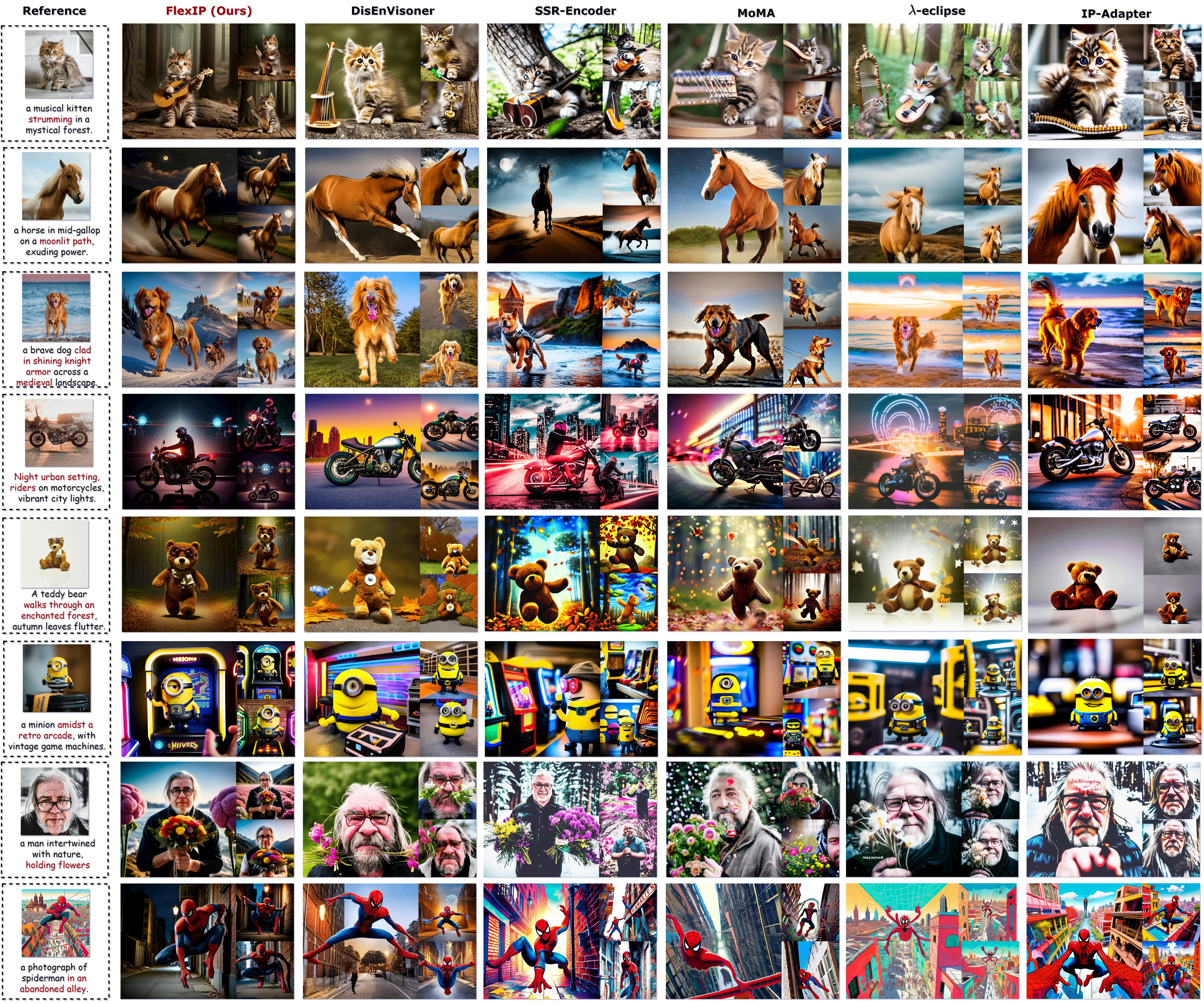}
    \caption{\textbf{Qualitative comparison with other methods}. Our approach surpasses alternative methods in its exceptional ability to preserve identity while generating a wide range of diverse and personalized outputs.}
    \label{fig:comp-main}
\end{figure*}

\subsection{Comparisons}

\subsubsection{Quantitative comparison}

In this experiment, we compared various methods in terms of personalization, preservation, image quality, and user study. The results are shown in the Tab.~\ref{tab:comparison}, where FlexIP outperformed all other methods across all evaluation metrics, particularly in mRank, personalization (CLIP-T), preservation (CLIP-I and DINO-I), image quality (CLIP-IQA and Aesthetic).
In terms of personalization, FlexIP scored 0.284 on CLIP-T, which is slightly lower than \(\lambda\)-Eclipse. However, \(\lambda\)-Eclipse achieves this at the expense of subject preservation abilities.  For preservation, FlexIP achieved high scores of 0.873 and 0.739 on CLIP-I and DINO-I, respectively, demonstrating its advantage in maintaining image details and semantic consistency. In image quality evaluation, FlexIP scored 0.598 on CLIP-IQA and 6.039 on Aesthetic, indicating superior quality and aesthetics of the generated images.


\begin{table}[ht]
\centering
\begin{tabular}{lccc}
\toprule
Method & I-T Match & Detail & Semantic \\
\midrule
\(\lambda\)-Eclipse    & 83.9 & 57.2 & 38.8 \\
DisEnVisioner & 66.6 & 56.9 & 38.6 \\
SSR-Encoder    & 83.1 & 56.1 & 38.5 \\
IP-Adapter    & 40.2 & 58.0 & 37.7 \\
MoMA   & 78.4 & 56.5 & 38.3 \\
\midrule
FlexIP    & \textbf{88.3} & \textbf{59.8} & \textbf{40.4} \\
\bottomrule
\end{tabular}
\caption{The evaluation metrics among different methods. Among these dimensions, I-T Match stands for image-text matching, Detail represents object detail satisfaction, and Semantic refers to semantic understanding. FlexIP surpasses previous methods across all three complementary indicators.}
\label{tab:text-following}
\end{table}

To provide a more human-aligned evaluation for personalization, we adopt the MLM-Filter~\cite{mlm-filter} to assess personalization. Unlike traditional methods like CLIP-T—which rely on global contrastive features and often miss fine-grained object details—The MLM-Filter utilizes advanced MLLM capabilities~\cite{liu2023llava,wang2024qwen2vl,bai2025qwen2.5vl} to capture subtle object properties and semantic nuances, enabling precise, context-aware evaluations aligned with human judgment.
Table~\ref{tab:text-following} demonstrates that FlexIP excels across three complementary dimensions—image-text matching (I-T Match), object detail satisfaction (Detail), and semantic understanding (Semantic). This highlights FlexIP’s ability to effectively capture subtle visual nuances and accurately integrate meaningful auxiliary information, closely aligning with human preferences and expectations.

To better demonstrate the effective of our methods, we evaluate the user satisfaction of different methods in practical applications, specifically focusing on flexibility (Flex) and identity preservation (ID-Pres). In this study, a total of 33 samples were utilized for evaluation purposes. During each evaluation session, participants were presented with a collection of images generated by various methods. A group of 60 evaluators was then asked to make selections based on two criteria: the image that best aligns with the textual semantics and the image that best preserves the subject. As show in Tab.~\ref{tab:comparison}, FlexIP excelled in both metrics.


\subsubsection{Qualitative comparison}
To further evaluate FlexIP’s capabilities, we present qualitative comparisons with five state-of-the-art methods across three distinct images per subject. As illustrated in Fig.~\ref{fig:comp-main}, FlexIP generates images with significantly enhanced fidelity, editability, and identity consistency compared to existing approaches. Fig.~\ref{fig:comp-main}  highlights FlexIP’s ability to maintain subject preservation and personalization across reference images under identical textual instructions, confirming the effectiveness of the explicit trade-off in the model.

\begin{figure}[ht]
    \centering
    \includegraphics[width=0.95\linewidth]{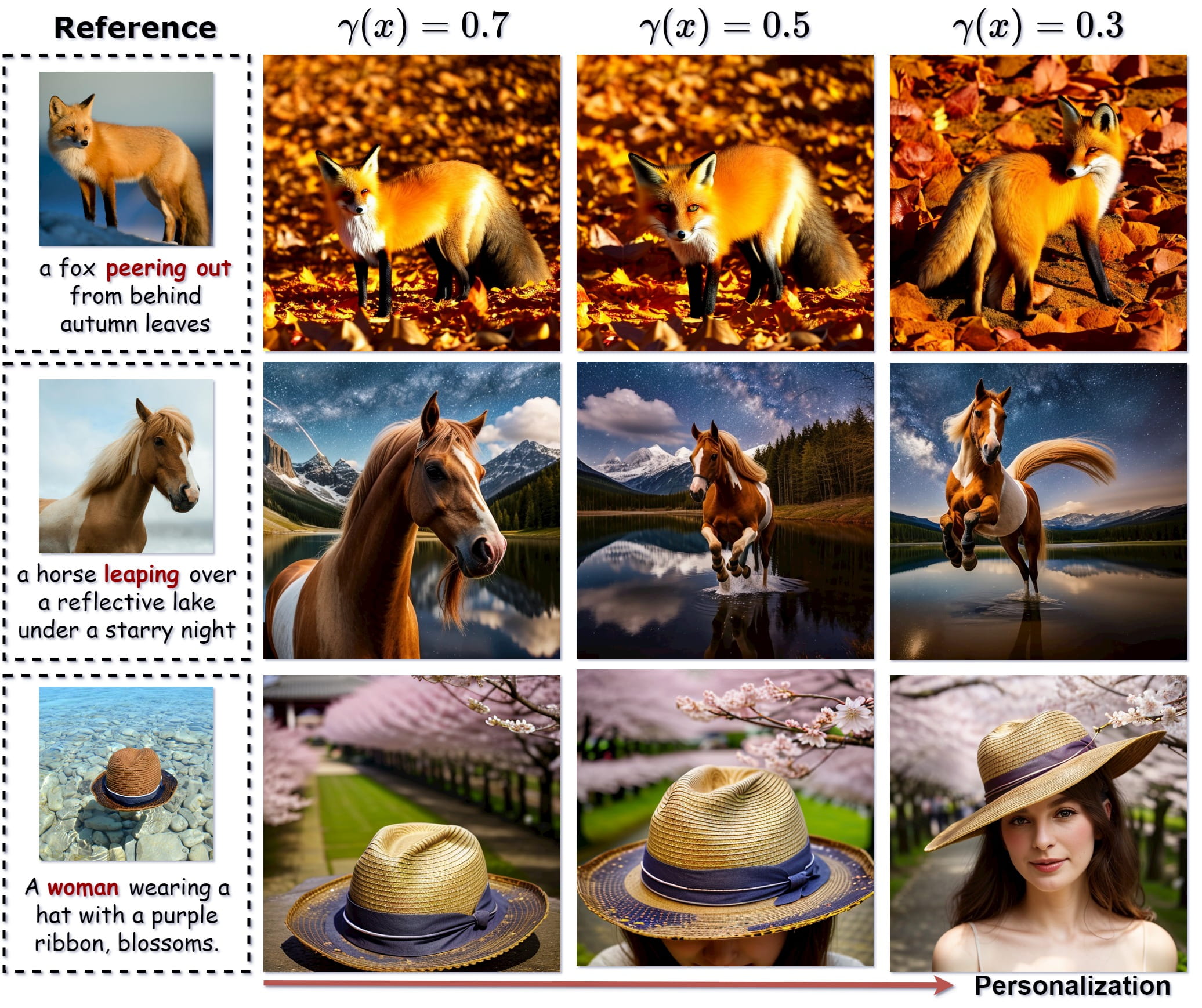}
    \caption{The effectiveness of the dynamic weight gating mechanism.}
    \label{fig:comp-var}
\end{figure}

\begin{figure}[ht]
    \centering
    \includegraphics[width=0.95\linewidth]{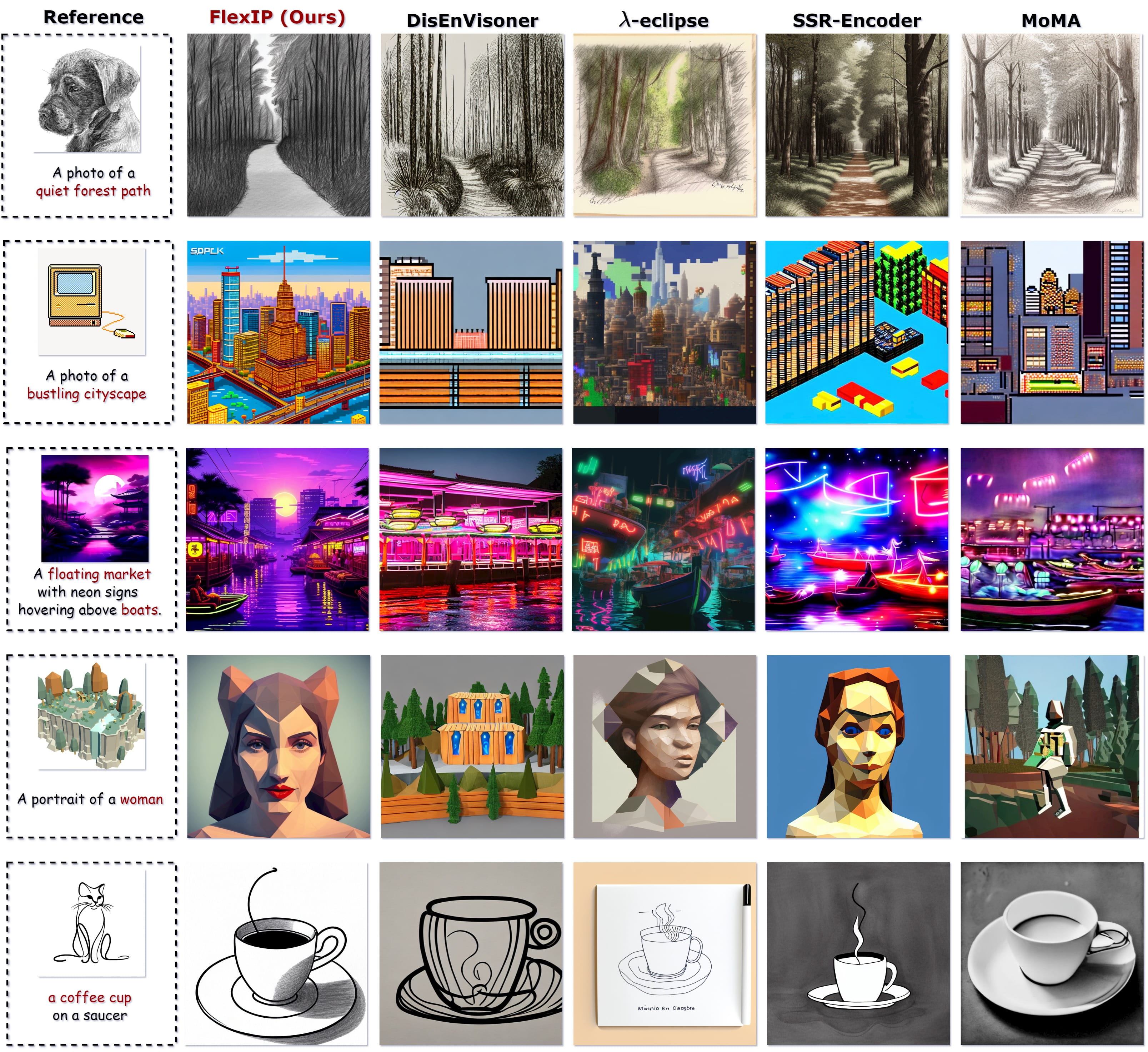}
    \caption{Comparison with other methods on style transfer tasks.}
    \label{fig:comp-style}
\end{figure}

\subsection{Ablation Study}

To validate the efficacy of the dynamic weight gating mechanism in explicitly balancing identity preservation and personalized editability, we conduct a comprehensive ablation study. As illustrated in Fig.~\ref{fig:comp-var}, our framework enables fine-grained control over the trade-off between these two objectives during inference by adjusting the relative weights of the preservation and personalization adapters. The proposed gating mechanism disentangles the optimization pathways of the two adapters during training, thereby mitigating the suboptimal performance caused by implicit trade-offs in joint optimization scenarios.

Qualitative results in Fig.~\ref{fig:comp-var} demonstrate that increasing the weight of the preservation adapter (e.g., $\gamma(x) \to 1$) prioritizes high-fidelity retention of the identity of the input subject, with minimal deviation in structural and textural details. In contrast, increasing the weight of the personalization adapter (e.g., $\gamma(x) \to 0$) improves editability, allowing greater stylistic transformations while maintaining semantic coherence. Critically, the linear interpolation between these weights enables users to smoothly traverse the preservation-editability spectrum at inference time, a capability absent in static fusion approaches. 

Furthermore, we extended the model to the task of zero-shot style transfer, emphasizing instruction following and detailed image information extraction. As demonstrated in Fig.~\ref{fig:comp-style}, our approach outperforms other methods in this task. This success is attributed to our dual adapter's ability to extract detailed information and maintain a balanced integration of detail extraction and instruction editing.

\section{Conclusion}

FlexIP is a novel framework for flexible subject attribute editing in image synthesis, effectively balancing identity preservation and personalized editing. By decoupling these objectives into independently controllable dimensions, FlexIP overcomes the limitations of existing methods. Its dual-adapter architecture ensures the maintenance of identity integrity by utilizing high-level semantic concepts and low-level spatial details. The dynamic weight gating mechanism allows users to control the trade-off between identity preservation and stylistic personalization, transforming the binary preservation-edit trade-off into a continuous parametric control surface which offer a robust and flexible solution for subject-driven image generation. 


{
    \small
    \bibliographystyle{ieeenat_fullname}
    \bibliography{main}
}
\clearpage
\appendix
\section{More Analysis}

\begin{figure}[ht]
    \centering
    \includegraphics[width=0.95\linewidth]{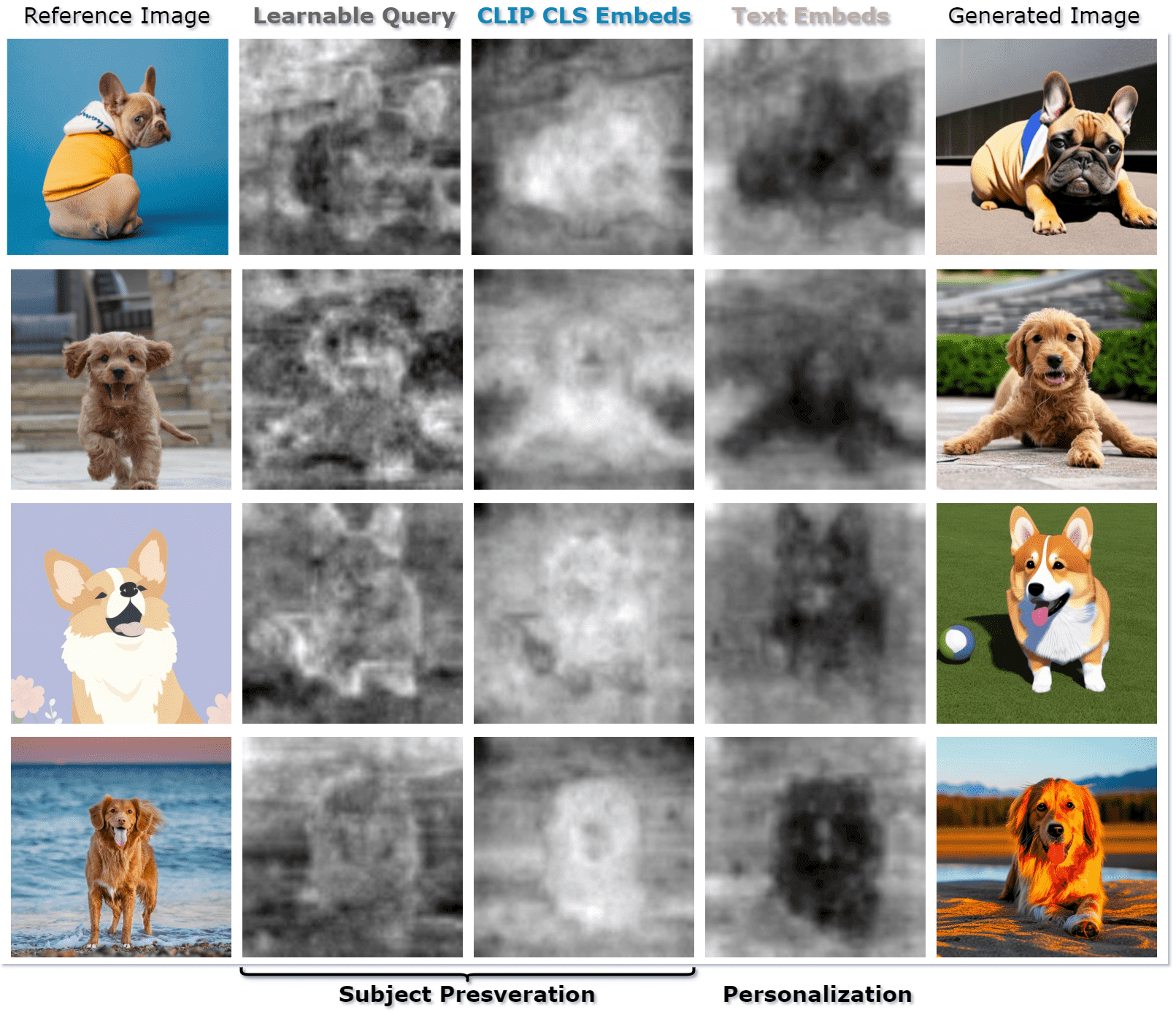}
    \caption{\textbf{Visualization of attention maps across different modules.}In the image, the white areas of the attention map indicate activation values—the whiter the color, the higher the activation value. It is evident that the two preservation modules function differently: the learnable query module concentrates more on the subject's details, while the CLIP CLS Embeds focus more on the subject's global aspects. Consequently, high-level and low-level information complement each other. For the personalization module, the text embeds pay more attention to the surrounding environment and some identity preservation details. This observation supports our decision to decouple preservation and personalization.}
    \label{fig:attnmaps-queries}
\end{figure}

As shown in Fig.~\ref{fig:attnmaps-queries}, we found that learnable queries specialize in capturing fine-grained variations but lack strong global coherence, while CLIP CLS embeddings provide global identity consistency but may miss subtle subject details. Therefore, instead of relying on a single embedding to learn both, we adopt a “divide and conquer” strategy that integrating both for retrieving fine-grained adaptability and global robustness simultaneously from DINO patch embeddings

\begin{table}[htbp]
    \centering
    \caption{\textbf{Data Information} used for training. Quality specifically refers to the image resolution.}
    \label{tab:data-information}

    \begin{tabular}{llcc}
        \toprule
        \textbf{Type} & \textbf{Dataset} & \textbf{Instances} & \textbf{Quality} \\
        \midrule
        \multicolumn{4}{c}{\textbf{Invariant Datasets} (11.1M) } \\
        \midrule
        \multirow{2}{*}{Image} 
        & SAM~\cite{kirillov2023segment}     & 9.0M   & High   \\
        & BrushData~\cite{ju2024brushnet} & 2.1M & Medium \\
        \midrule
        \multicolumn{4}{c}{\textbf{Variant Datasets} (1.23M) } \\
        \midrule
        \multirow{8}{*}{Multi-View} 
            & MVImageNet~\cite{yu2023mvimgnet} & 177495 & Medium \\
            & MVHumanNet~\cite{xiong2024mvhumannet} & 28893            & High   \\
            & co3d~\cite{reizenstein2021common}       & 26687            & Low    \\
            & PanoHead~\cite{an2023panohead}   & 5000 & Medium \\
            & CelebA~\cite{liu2015faceattributes} & 10133  & High \\
            & MeGlass~\cite{guo2018face}    & 1710 & Low    \\
            & VITON-HD~\cite{choi2021viton}   & 11647            & High   \\
            & DressCode~\cite{morelli2022dress}  & 53792            & Medium \\
        \midrule
        \multirow{5}{*}{Video} 
            & SAM2~\cite{ravi2024sam}       & 51000  & High   \\
            & CelebV-HQ~\cite{zhu2022celebv}  & 35666  & Medium \\
            & VFHQ~\cite{xie2022vfhq}       & 15204  & Medium \\
            & Pexel      & 181038 & High   \\
            & OpenVid1M~\cite{nan2025openvidm}  & 633885 & High   \\
        \bottomrule
    \end{tabular}

\end{table}

\section{More Experimental Details}

\begin{figure*}[ht]
    \centering
    \includegraphics[width=1.0\linewidth]{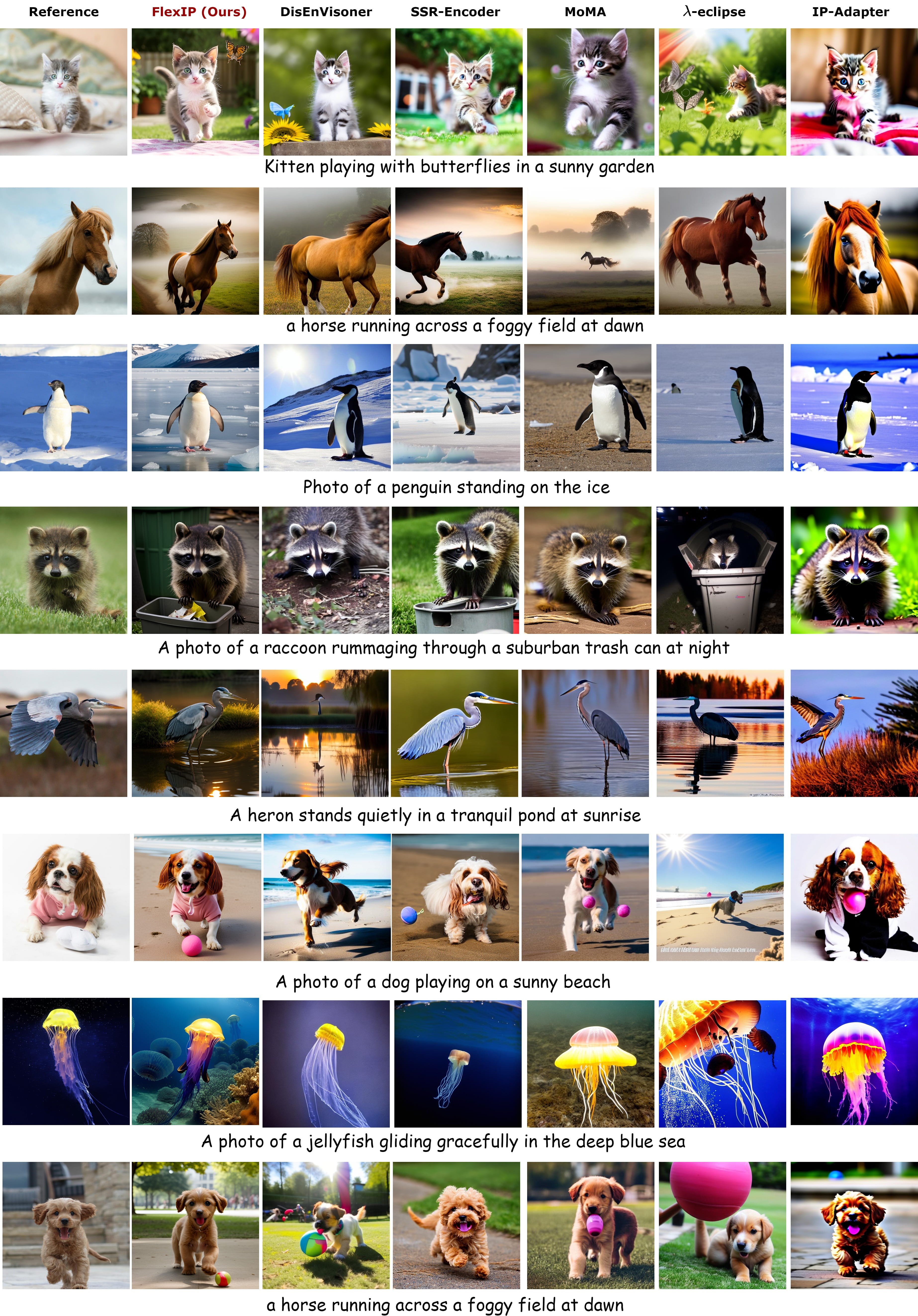}
    \caption{\textbf{Qualitative comparison with other methods in animal domain}. Our approach surpasses alternative methods in its exceptional ability to preserve identity while generating a wide range of diverse and personalized outputs.}
    \label{fig:supp-animals}
\end{figure*}

\begin{figure*}[ht]
    \centering
    \includegraphics[width=1.0\linewidth]{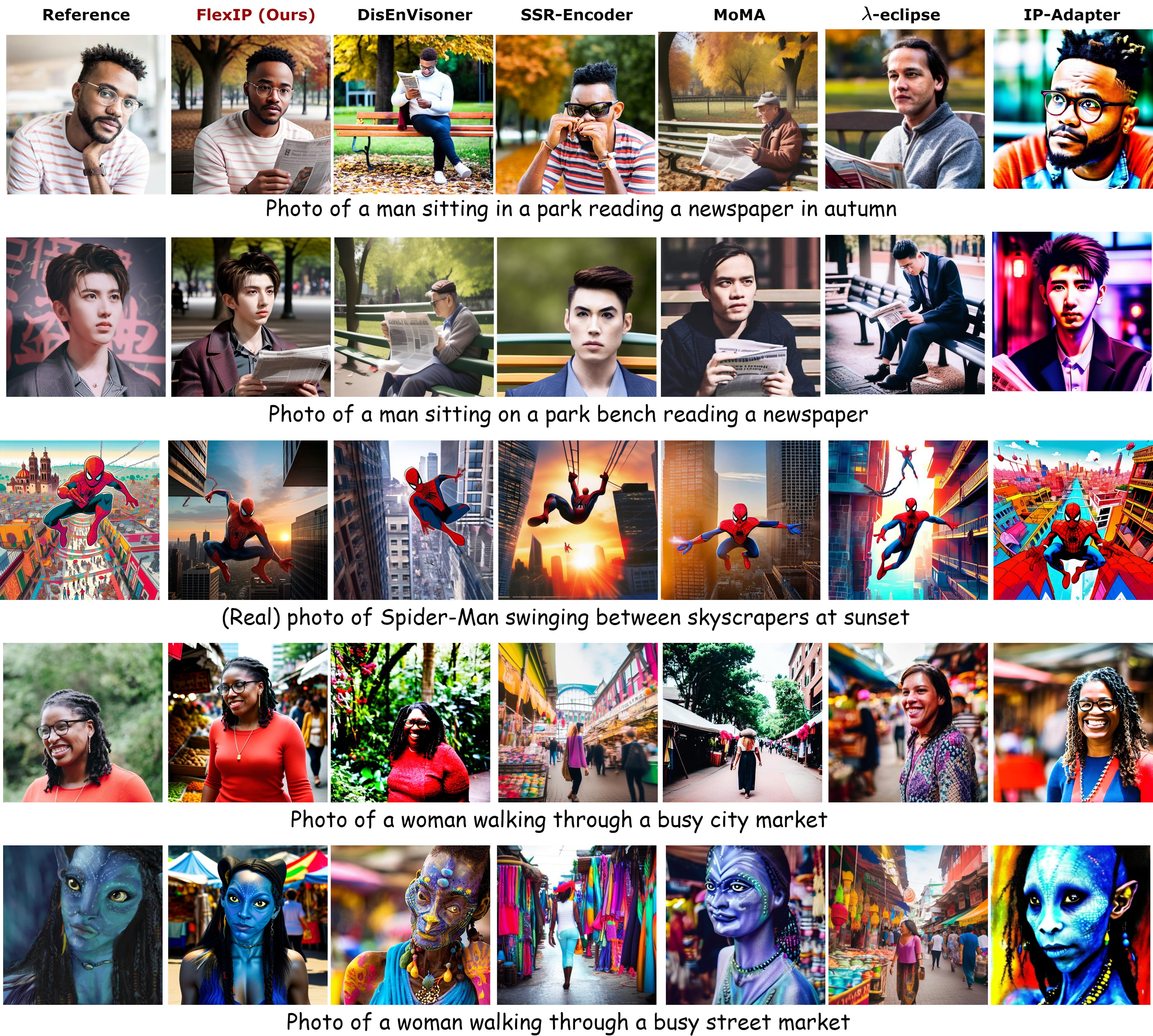}
    \caption{\textbf{Qualitative comparison with other methods in human domain}. Our approach surpasses alternative methods in its exceptional ability to preserve identity while generating a wide range of diverse and personalized outputs.}
    \label{fig:supp-human}
\end{figure*}

\begin{figure*}[ht]
    \centering
    \includegraphics[width=1.0\linewidth]{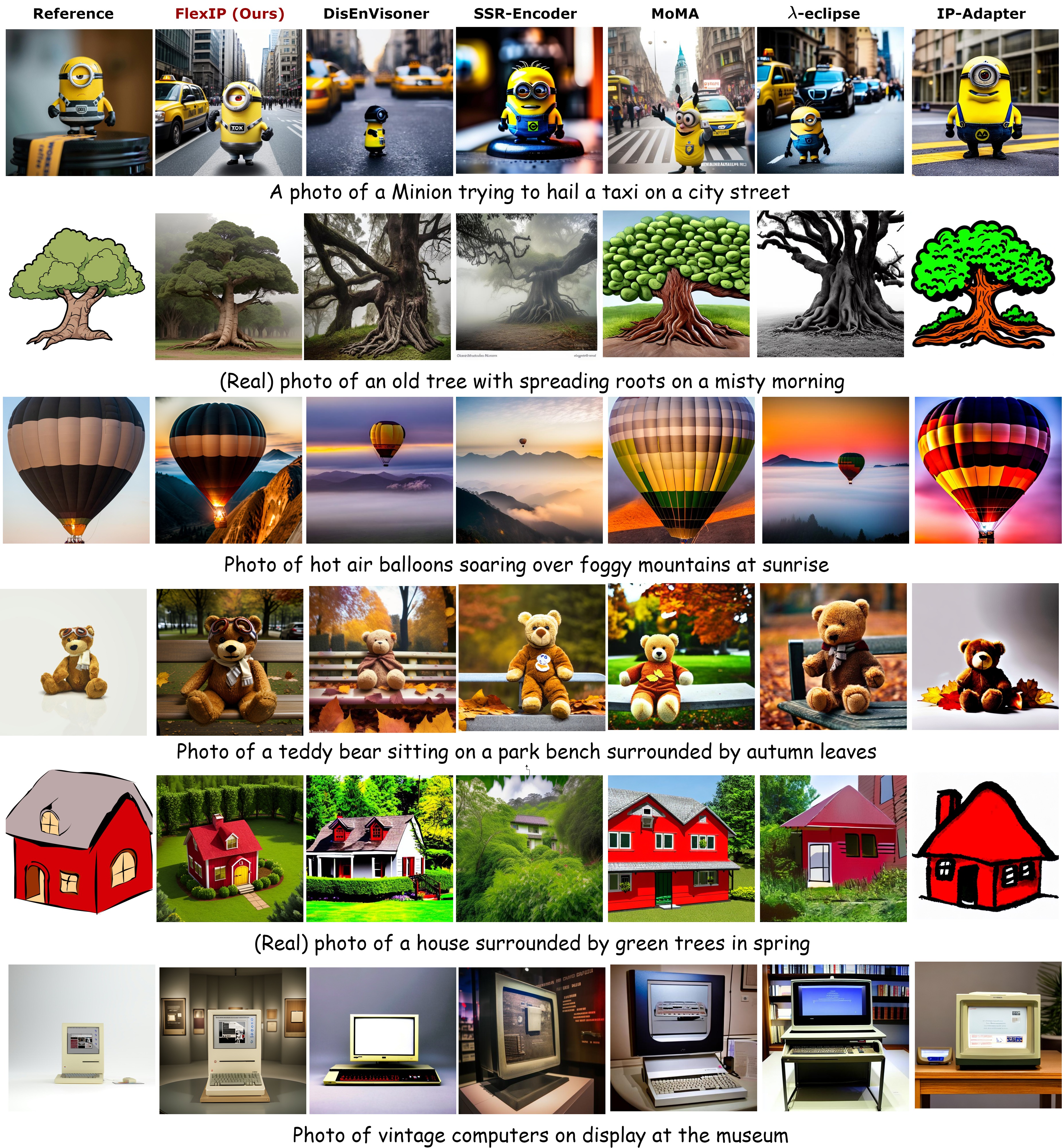}
    \caption{\textbf{Qualitative comparison with other methods in object domain}. Our approach surpasses alternative methods in its exceptional ability to preserve identity while generating a wide range of diverse and personalized outputs.}
    \label{fig:supp-object}
\end{figure*}

\subsection{Implementation Details}

FlexIP is built on Stable Diffusion v1.5, utilizing OpenCLIP ViT-H/14~\cite{radford2021learning} as the high-level image encoder and DinoV2-L~\cite{oquab2023dinov2} as the low-level image encoder. The model is trained on 8 GPUs with 32GB of memory for 140,000 steps at a resolution of 512×512, with a batch size of 16 per GPU, a learning rate of 1e-4, and a weight decay of 0.01. After 100,000 steps, it undergoes fine-tuning with higher-quality images for an additional 40,000 steps. During training, classifier-free guidance is applied with a 5\% probability of dropping text, images, or both. For inference, DDIM sampling with 50 steps and a guidance scale of 7.5 is used. 
As shown in Table~\ref{tab:data-information}, we include various types of datasets used for training.

\section{Background}

\textbf{Diffusion Models.} 
Diffusion models comprise a family of generative models characterized by two fundamental processes:
(i) a \textit{diffusion process} (forward process), which gradually corrupts data through a fixed $T$-step Markov chain by adding Gaussian noise, and
(ii) a \textit{denoising process} that iteratively recovers data from noise via a learnable model. For conditional variants like text-to-image models, the denoising process is guided by auxiliary inputs such as text prompts.

The training objective for the noise prediction network $\epsilon_\theta$ optimizes a simplified variational bound:
\begin{equation}
\mathcal{L}_{\text{simple}} = \mathbb{E}_{x_0, \epsilon \sim \mathcal{N}(0, \mathbf{I}), c, t} \left| \epsilon - \epsilon_\theta(x_t, c, t) \right|^2,
\end{equation}
where $x_0$ denotes clean data, $c$ represents conditioning signals, $t \in \{1, \ldots, T\}$ indexes the diffusion timestep, and $x_t = \alpha_t x_0 + \sigma_t \epsilon$ describes the noisy state at step $t$ with $\alpha_t, \sigma_t$ being predefined noise scheduling coefficients.

During inference, initial noise $x_T \sim \mathcal{N}(0, \mathbf{I})$ is progressively denoised through $T$ iterations. Accelerated sampling is typically achieved via deterministic ODE solvers like DDIM \citep{song2021denoising}, PNDM \citep{liu2022pseudo}, or adaptive-step methods like DPM-Solver \citep{lu2022dpm}.

For conditional diffusion models, \textit{classifier guidance} \citep{dhariwal2021diffusion} balances image fidelity and sample diversity by leveraging gradients from an independently trained classifier. To circumvent the requirement for a separate classifier, \textit{classifier-free guidance} \citep{ho2021classifierfree} is widely adopted. This method jointly trains conditional and unconditional denoising paths by randomly omitting the condition $c$ with probability $p_{\text{drop}}$ during training. At inference, the noise prediction is interpolated between the conditional and unconditional outputs:

\begin{equation}
\hat{\epsilon}_\theta(x_t, c, t) = w \cdot \epsilon_\theta(x_t, c, t) + (1 - w) \cdot \epsilon_\theta(x_t, t),
 \end{equation}

where $w > 1$ (termed the \textit{guidance scale}) amplifies alignment with the condition $c$. For text-to-image diffusion models, this mechanism critically strengthens the semantic correspondence between generated images and text prompts.

In our work, we implement the \textit{FlexIP} atop the open-source \textit{Stable Diffusion (SD)} framework \citep{rombach2022high}. SD operates as a latent diffusion model conditioned on text embeddings from a frozen CLIP text encoder \citep{radford2021learning}. Its backbone comprises a time-conditional U-Net \citep{ronneberger2015u} with cross-attention layers that fuse text features into the diffusion process. Unlike pixel-space models (e.g., Imagen \citep{saharia2022photorealistic}), SD achieves computational efficiency by performing diffusion in the latent space of a pretrained variational autoencoder, reducing dimensionality by a factor of 4–64 compared to raw pixels.

\end{document}